\documentclass[conference]{IEEEtran}
\IEEEoverridecommandlockouts
\usepackage{cite}
\usepackage{amsmath,amssymb,amsfonts}
\usepackage{algorithmic}
\usepackage{graphicx}
\usepackage{textcomp}
\usepackage{xcolor}
\usepackage{comment}
\usepackage{float}
\usepackage{algorithm}
\usepackage{multirow}
\usepackage{hyperref}

\def\BibTeX{{\rm B\kern-.05em{\sc i\kern-.025em b}\kern-.08em
    T\kern-.1667em\lower.7ex\hbox{E}\kern-.125emX}}
\begin{document}

\makeatletter
\def\ps@IEEEtitlepagestyle{%
  \def\@oddfoot{\mycopyrightnotice}%
  \def\@evenfoot{}%
}
\def\mycopyrightnotice{%
  {\footnotesize
    \parbox[b]{\textwidth}{%
      © 2026 IEEE. Personal use of this material is permitted.
      Permission from IEEE must be obtained for all other uses, in any current
      or future media, including reprinting/republishing this material for
      advertising or promotional purposes, creating new collective works, for
      resale or redistribution to servers or lists, or reuse of any copyrighted
      component of this work in other works.%
    }%
  }%
  \gdef\mycopyrightnotice{}%
}
\makeatother

\title{A Generative AI Integrated Multimodal Framework for Low-Latency Multi-Camera Person Re-Identification
}

\author{
Leon Fernando$^{1}$, C Dombawala$^{2}$, P. Hettigoda$^{2}$, VG~Warnasooriya$^{3}$, Ishara Neranjana$^{4}$, Rashmika Nawaratne$^{5}$\\
\small
\parbox{\textwidth}{
$^{1}$University of Moratuwa, 
$^{2}$Zone24x7 (Pvt) Ltd,  
$^{3}$Chulalongkorn University,
$^{4}$University of Colombo, 
$^{5}$La Trobe University, Australia
}
}

\maketitle

\begin{abstract}
Person re-identification (ReID) is essential for multi-camera surveillance and tracking, yet remains difficult due to viewpoint and illumination changes, occlusion, background clutter, and low resolution imagery. We propose a generative AI integrated multimodal ReID framework designed explicitly for robustness under missing cues and low latency deployment. The key idea is a cost aware early-exit cascade that prioritizes inexpensive, high confidence evidence and only triggers expensive modalities for ambiguous cases.

Our system integrates (i) global visual embeddings from segmented person regions, (ii) automatically generated fine grained semantic attribute descriptions generated by vision-language models (VLMs), and (iii) optional facial embeddings when face observations are reliable. To optimize the balance between accuracy and latency, we use a cost aware early-exit cascade instead of fusing all modalities. Specifically, we first inspect the top-$k$ retrieval results to determine whether the query is unambiguous. If the best match is clearly separated from the remaining candidates, we stop early and return the result to minimize latency; in ambiguous cases, we keep multiple hypotheses and invoke additional modalities (face/semantic) with adaptive reliability weighting to refine the decision.

We report person re-identification performance using mAP and Rank-1 accuracy on the Market-1501 and DukeMTMC-reID benchmarks. The proposed adaptive early-exit cascade resolves 60.7\% of DukeMTMC-reID queries and 68.4\% of Market-1501 queries without invoking semantic reasoning, reducing computational overhead while maintaining competitive retrieval performance.

\textit{GitHub: \url{https://github.com/leonfdo/DeepReIDers/tree/develop}}

\end{abstract}


\section{\textbf{Introduction}}

Person re-identification (ReID) aims to match pedestrian identities across different camera views and time, and is a key component in intelligent surveillance, forensic retrieval, and multi-camera tracking. With the rapid growth of large scale camera networks, automated ReID systems have become essential for scalable real world monitoring. Despite significant progress, deploying ReID systems in real world environments remains challenging. Variations in viewpoint, illumination, occlusion, motion blur, and background clutter affect feature consistency across cameras. In addition, practical scenarios often involve low resolution data and environmental noise, which degrade feature reliability. These systems must also operate under limited computational resources, making it difficult to balance accuracy with real time performance.

Recent advances in deep learning, especially vision-language pretraining, have improved representation learning by aligning visual and semantic information \cite{clip}. This enables better generalization across diverse environments. However, many high performing ReID methods focus primarily on accuracy, often overlooking computational efficiency, which limits their use in latency sensitive applications. In real world settings, the availability of modalities such as facial features or semantic attributes is dynamic. Faces may be occluded or absent, and semantic cues may vary depending on conditions. Existing multimodal ReID systems typically use an always on fusion approach, processing all modalities for every query. This leads to unnecessary computational overhead and increased inference latency, even in simple scenarios.

To address these limitations, we propose a generative AI integrated multimodal ReID framework designed for both robustness and efficiency. The proposed system introduces a cost aware early-exit cascade mechanism that adaptively determines the level of computation required for each query. Instead of uniformly applying all modalities, the system first evaluates inexpensive and high confidence visual cues, and selectively incorporates additional modalities such as semantic embeddings generated by VLMs and facial features only when necessary. Furthermore, a reliability aware fusion strategy dynamically adjusts modality contributions based on their availability and confidence. The main contributions of this paper are as follows:

\begin{enumerate}
\renewcommand{\labelenumi}{\arabic{enumi})}
    \item A cost aware early-exit cascade mechanism for multimodal person re-identification that adaptively reduces unnecessary computation while maintaining high identification accuracy.
    \item Integration of generative AI based semantic embeddings with visual and facial features to improve robustness under challenging conditions such as occlusion and domain variability.
    \item A reliability aware multimodal fusion strategy that dynamically adjusts modality contributions based on their availability and confidence.
\end{enumerate}

\section{\textbf{Related Work}}
\label{sec:related_work}

Recent advancements in person re-identification (ReID) have focused on improving robustness through semantic understanding, multimodal learning, and efficient inference. Vision-language pretraining has emerged as a powerful paradigm, where models align visual and textual representations to capture high-level contextual attributes beyond raw appearance~\cite{clip}. CLIP-based approaches demonstrate strong performance gains by incorporating semantic information into ReID pipelines~\cite{clipreid}. Subsequent works extend this idea using synthesized captions and large VLMs to further enhance representation quality in visually ambiguous scenarios~\cite{clipscgi, lvlmreid}. 

Despite these advances, domain discrepancies between training and deployment environments remain a major challenge. Variations in illumination, background, and camera characteristics can significantly degrade performance. Domain adaptation methods such as generative transfer aim to bridge this gap~\cite{domaingap}, while recent studies highlight persistent limitations of CLIP based approaches under domain shifts~\cite{wang2025visualanalysis}.

In parallel, strong visual backbones such as OSNet x1.0 enable efficient omni scale feature learning for robust appearance representation~\cite{osnet}. Multimodal ReID further improves performance by combining complementary cues, including semantic features and auxiliary biometrics such as facial embeddings. Face recognition models such as MTCNN and FaceNet provide discriminative identity cues when reliable facial observations are available~\cite{mtcnn,facenet}. However, most existing systems rely on fixed or always on fusion strategies, where all modalities are processed for every query, leading to unnecessary computational overhead.

To improve efficiency, prior work has explored staged and coarse to fine retrieval strategies, where lightweight representations are used to filter candidates before applying more expensive matching~\cite{fasterreid}. In addition, real time detection frameworks such as YOLO enable efficient person localization in large scale systems~\cite{yolo}. Nevertheless, most existing approaches employ static inference pipelines that do not adapt to query difficulty. In practice, many queries can be resolved using inexpensive features, while only ambiguous cases require additional processing.

Video based ReID further enhances robustness by leveraging temporal information across multiple frames, reducing noise and improving identity consistency under occlusion and motion variations~\cite{videoreid}. However, uniform processing of all frames introduces redundancy and increases computational cost, highlighting the need for selective and efficient aggregation strategies.

Although frameworks such as FastReID provide scalable and optimized ReID pipelines~\cite{fastreid}, most prior work focuses on either accuracy or efficiency independently. In contrast, our approach combines vision-language reasoning, visual features, and facial cues within an adaptive early-exit cascade framework, enabling selective computation and improving both robustness and efficiency under dynamic conditions.

\section{\textbf{Methodology}}

\begin{figure}[H]
    \centering
    \includegraphics[width=\columnwidth]{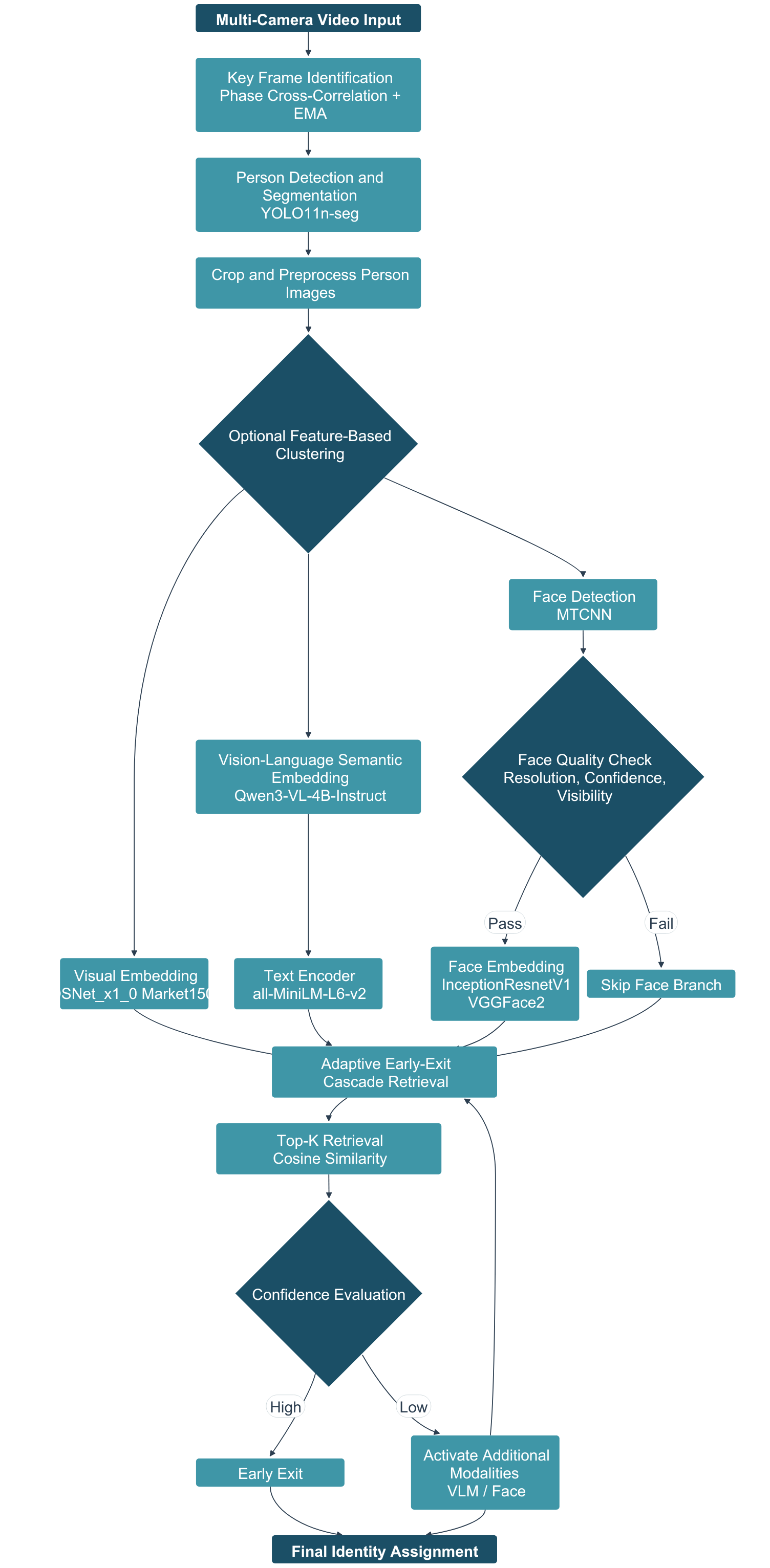}
    \caption{Proposed generative AI integrated multimodal framework for multi-camera person re-identification with early-exit cascade retrieval and temporal aggregation.}
    \label{fig:methodology}
\end{figure}

As illustrated in Fig.~\ref{fig:methodology}, the proposed framework introduces an adaptive multimodal architecture for multi-camera person re-identification (ReID), designed to achieve robust identification under real world conditions while maintaining low latency performance. 

\begin{algorithm}[h]
\caption{Gallery Construction with Adaptive Cascade ReID}
\label{alg:gallery}
\begin{algorithmic}[1]

\STATE $G \leftarrow \emptyset$ \COMMENT{Initialize empty gallery}
\STATE $\text{seen\_ids} \leftarrow \emptyset$ \COMMENT{Track enrolled person IDs}

\FOR{each image $I_i \in \mathcal{I}$}

    \STATE \textbf{// Step 1: Detection \& Embedding Extraction}
    \STATE $\text{crops} \leftarrow \text{DETECT\_PERSONS}(I_i)$
    
    \IF{$\text{crops} = \emptyset$}
        \STATE \textbf{continue}
    \ENDIF
    
    \STATE $\text{crop} \leftarrow \text{SELECT\_LARGEST}(\text{crops})$
    \STATE $q_{\text{vis}} \leftarrow \text{EXTRACT\_VISUAL}(\text{crop})$
    \STATE $q_{\text{sem}} \leftarrow \text{EXTRACT\_SEMANTIC}(\text{crop})$ \COMMENT{NULL if disabled}
    \STATE $q_{\text{face}} \leftarrow \text{EXTRACT\_FACE}(\text{crop})$ \COMMENT{NULL if no face}

    \STATE \textbf{// Step 2: Cascade Retrieval}
    \STATE $R \leftarrow \text{CASCADE\_SEARCH}(G,\; q = (q_{\text{vis}}, q_{\text{sem}}, q_{\text{face}}))$

    \IF{$R \neq \emptyset$}
        \STATE \textbf{// Match found}
        \STATE $\text{best} \leftarrow R[1]$
        \STATE $\text{id}_{\text{matched}} \leftarrow \text{best.person\_id}$
        
    \ELSE
        \STATE \textbf{// New identity enrollment}
        \STATE $\text{id}_{\text{new}} \leftarrow \text{EXTRACT\_ID}(I_i)$
        \STATE $\text{UPSERT}(G,\; \text{id}_{\text{new}},\; q_{\text{vis}},\; q_{\text{sem}},\; q_{\text{face}})$
        \STATE $\text{seen\_ids} \leftarrow \text{seen\_ids} \cup \{\text{id}_{\text{new}}\}$
    \ENDIF

\ENDFOR

\end{algorithmic}
\end{algorithm}

The overall pipeline consists of person detection and segmentation, multimodal feature extraction, adaptive cascade based retrieval, and temporal aggregation, guided by a cost aware processing strategy for efficient real time deployment. 

Video streams from multiple cameras are first processed using a real time detection and segmentation model (YOLO11n-seg) to localize person instances. Segmentation masks refine detected regions by isolating foreground content, reducing background interference. The resulting cropped person regions are then forwarded to the feature extraction stage.

Visual features are extracted using OSNet x1.0 pretrained on Market-1501, providing discriminative appearance representations for efficient retrieval. Semantic representations are generated using the VLM Qwen3-VL-4B-Instruct, which produces fine grained textual descriptions of person attributes. These descriptions are then encoded using all-MiniLM-L6-v2 to obtain compact semantic embeddings. This semantic branch captures high-level identity cues such as clothing attributes and contextual appearance, significantly improving robustness under viewpoint variations, occlusion, and low-resolution conditions.

Additionally, facial information is incorporated when available. Face regions are detected using MTCNN, and a quality aware gating mechanism evaluates their reliability based on resolution, visibility, and detection confidence. Only valid face regions are processed using an InceptionResNetV1 model pretrained on VGGFace2 to extract discriminative facial embeddings, avoiding the influence of low quality inputs.

To balance accuracy and computational efficiency, an adaptive early-exit cascade retrieval mechanism is employed. 

\begin{algorithm}[H]
\caption{CASCADE\_SEARCH: Multi-Modal Cascade Retrieval}
\label{alg:cascade}
\begin{algorithmic}[1]

\STATE \textbf{Input:} $G$, $q=(q_{\text{vis}}, q_{\text{sem}}, q_{\text{face}})$
\STATE \hspace{1.2cm} $K_1, K_2, K_3, K_{\text{final}}$
\STATE \hspace{1.2cm} $t_{\text{exit}}, t_{\text{esc}}, t_{\text{nf}}, d_{\text{nf}}, t_{\min}$
\STATE \hspace{1.2cm} $w_{\text{vis}}, w_{\text{sem}}, w_{\text{face}}$
\STATE \textbf{Output:} $R$

\vspace{0.3em}
\STATE \textbf{// Stage 1: Visual Retrieval}
\STATE $C_1 \leftarrow \text{SEARCH}(G_{\text{visual}}, q_{\text{vis}}, K_1)$
\STATE $s_1 \leftarrow \text{SCORES}(C_1)$

\IF{$s_1 = \emptyset$ \OR $\max(s_1) < t_{\min}$}
    \STATE \textbf{return} $\emptyset$
\ENDIF

\STATE $d_1 \leftarrow \text{QUALITY\_EVALUATE}(s_1, \text{stage}=1)$

\IF{$d_1 = \text{EARLY\_EXIT}$}
    \STATE \textbf{return} $\text{TOP}(C_1, K_{\text{final}})$
\ELSIF{$d_1 = \text{NO\_MATCH}$}
    \STATE \textbf{return} $\emptyset$
\ENDIF

\vspace{0.3em}
\STATE \textbf{// Stage 2: Semantic Refinement}

\IF{$q_{\text{sem}} = \text{NULL}$}
    \STATE $S_{\text{fused}} \leftarrow s_1$
    \STATE $C_{\text{fused}} \leftarrow C_1$
\ELSE
    \STATE $C_2 \leftarrow \text{SEARCH}(G_{\text{semantic}}, q_{\text{sem}}, K_2)$
    \STATE $S_{\text{fused}} \leftarrow \text{FUSE}(s_1,\; \text{SCORES}(C_2),\; w_{\text{vis}}, w_{\text{sem}})$
    \STATE $C_{\text{fused}} \leftarrow \text{SORT\_BY}(C_1 \cup C_2,\; S_{\text{fused}})$

    \STATE $d_2 \leftarrow \text{QUALITY\_EVALUATE}(S_{\text{fused}}, \text{stage}=2)$

    \IF{$d_2 = \text{EARLY\_EXIT}$}
        \STATE \textbf{return} $\text{TOP}(C_{\text{fused}}, K_{\text{final}})$
    \ELSIF{$d_2 = \text{NO\_MATCH}$}
        \STATE \textbf{return} $\emptyset$
    \ENDIF
\ENDIF

\vspace{0.3em}
\STATE \textbf{// Stage 3: Face Verification}

\IF{$q_{\text{face}} \neq \text{NULL}$}
    \STATE $C_3 \leftarrow \text{SEARCH}(G_{\text{face}}, q_{\text{face}}, K_3)$
    \STATE $S_{\text{final}} \leftarrow \text{FUSE\_THREE}($
    \STATE \hspace{2.2em} $s_1,\; \text{SCORES}(C_2),\; \text{SCORES}(C_3),$
    \STATE \hspace{2.2em} $w_{\text{vis}},\; w_{\text{sem}},\; w_{\text{face}})$
    \STATE $S_{\text{final}} \leftarrow \text{SUPPRESS}(S_{\text{final}})$
    \STATE \textbf{return} $\text{TOP}(\text{SORT\_BY}(C_{\text{fused}}, S_{\text{final}}), K_{\text{final}})$

\ELSE
    \STATE \textbf{// No-face decision gate}
    \STATE $\text{margin} \leftarrow S_{\text{fused}}[1] - S_{\text{fused}}[2]$

    \IF{$\max(S_{\text{fused}}) < t_{\text{nf}}$ \OR $\text{margin} < d_{\text{nf}}$}
        \STATE \textbf{return} $\emptyset$
    \ENDIF

    \STATE \textbf{return} $\text{TOP}(C_{\text{fused}}, K_{\text{final}})$
\ENDIF

\end{algorithmic}
\end{algorithm}

Retrieval is initially performed using visual embeddings, where top-$k$ candidates are obtained based on cosine similarity. A confidence evaluation module determines whether the top match is sufficiently reliable. If the confidence exceeds a predefined threshold, the system performs an early-exit, reducing inference latency. Otherwise, additional modalities are dynamically activated, and semantic embeddings are selectively invoked only for ambiguous cases, followed by optional facial verification. Final identity prediction is obtained through adaptive fusion based on modality availability and reliability.

A quality aware filtering stage further refines similarity scores by suppressing ambiguous matches using statistical measures such as inter class similarity margins, improving robustness in complex environments. For video based ReID, a temporal aggregation module stabilizes predictions across frames. Low quality frames are filtered based on detection confidence and motion consistency, while the remaining embeddings are aggregated using mean pooling or exponential moving average (EMA) to produce stable track level representations.

Algorithm~\ref{alg:gallery} describes the gallery construction and multimodal feature extraction procedure, while Algorithm~\ref{alg:cascade} summarizes the adaptive cascade retrieval and early-exit strategy used during inference. Together, these components enable the framework to achieve a balance between accuracy and computational efficiency through adaptive early-exit retrieval, selective modality activation, and efficient temporal aggregation, making it suitable for real time multi-camera deployment.

\section{\textbf{Experimental Evaluation}}
\label{sec:experiments}

\subsection{\textbf{Experimental Settings}}
\label{sec:setup}

\subsubsection{\textbf{Datasets and Implementation Details}}

We evaluate the proposed framework on two widely used person re-identification benchmark ReID datasets. DukeMTMC-reID~\cite{duke} contains 17,661 gallery images of 1,110 identities and 2,228 query images of 702 identities captured across 8 camera views. Market-1501~\cite{market} consists of 13,115 gallery images of 750 identities and 3,368 query images of 750 identities captured from 6 cameras. 

The proposed framework is implemented in PyTorch and executed on a GPU-enabled system equipped with an NVIDIA RTX 5070 Ti Laptop GPU (12\,GB VRAM, SM 12.0, Blackwell architecture) using CUDA 12.8 (PyTorch 2.11.0+cu128), achieving approximately $\sim$84\% GPU utilization during execution.

\subsubsection{\textbf{Evaluation Metrics}}
We adopt standard person re-identification evaluation metrics, including Cumulative Matching Characteristics (CMC) at Rank-1, Rank-5, and Rank-10, and mean Average Precision (mAP), to provide a comprehensive assessment of retrieval accuracy under different ranking conditions.

\subsection{\textbf{Quantitative Results}}
\subsubsection{\textbf{Overall Performance}}

Table~\ref{tab:compact_sota} presents a comparison between the proposed framework and existing CNN-, ViT-, and LVLM-based ReID methods. While LVLM-based approaches achieve strong performance due to their rich semantic understanding, they require invoking large VLMs for every query, resulting in significant computational overhead and increased inference time.

\label{sec:quantitative}

\begin{table}[H]
\centering
\caption{Comparison with State-of-the-Art Methods}
\label{tab:compact_sota}
\small
\setlength{\tabcolsep}{2.5pt}
\renewcommand{\arraystretch}{1.05}

\begin{tabular}{llcc|cc}
\hline
\textbf{Backbone} & \textbf{Method} 
& \multicolumn{2}{c|}{\textbf{DukeMTMC-reID}} 
& \multicolumn{2}{c}{\textbf{Market-1501}} \\
\cline{3-6}
 &  & \textbf{mAP} & \textbf{Rank-1} & \textbf{mAP} & \textbf{Rank-1} \\
\hline

\multirow{4}{*}{CNN}
& MGN~\cite{mgn}        & 78.4  & 88.7 & 86.9 & \textbf{95.7} \\
& DG-Net~\cite{dgnet}   & 74.8  & 86.6 & 86.0 & 94.8 \\
& SAN~\cite{san}        & 75.5  & 87.9 & 88.0 & 96.1 \\
& Pyramid~\cite{pyramid}& 79.0  & 89.0 & 88.2 & \textbf{95.7} \\
\hline

\multirow{3}{*}{ViT}
& TransReID~\cite{transreid} & 80.6 & 89.6 & 88.2 & 95.0 \\
& PFD~\cite{pfd}             & 82.2 & 90.6 & \textbf{89.6} & 95.5 \\
& CLIP-ReID~\cite{clipreid}  & 82.5 & 90.0 & \textbf{89.6} & 95.5 \\
\hline

LVLM
& LVLM-ReID~\cite{lvlmreid} & \textbf{82.8} & \textbf{92.2} & 89.2 & 95.6 \\
\hline

Multimodal
& DeepReID(Ours) & 81.9 & 90.5 & 88.9 & 95.2 \\
\hline

\end{tabular}
\end{table}

Similarly, CNN and ViT-based methods demonstrate competitive accuracy; however, they rely heavily on supervised training with dataset-specific features. This makes them less adaptable to unseen environments and limits their generalization capability without extensive retraining.

In contrast, the proposed framework adopts a different design philosophy by leveraging pretrained models combined with prompt-optimized semantic representations. This eliminates the need for costly task specific training while maintaining reasonable performance across benchmarks. More importantly, the system is explicitly designed for low-latency operation, where the primary objective is not to outperform all methods in accuracy, but to reduce computational cost and inference time while preserving acceptable accuracy levels.

By integrating an adaptive early-exit cascade mechanism, the framework avoids unnecessary activation of expensive modalities (such as VLMs) for every query. This enables efficient processing, making the system more suitable for real time, large scale multi-camera deployments where latency is a critical constraint.

\subsubsection{\textbf{Latency and Efficiency Analysis}}

Table~\ref{tab:latency} presents the component-wise latency analysis of the proposed framework. Detection + Crop, Cascade Retrieval, and Vector DB Update are executed for every query, forming a lightweight baseline computational cost with minimal overhead.

The total inference time varies depending on the execution path selected by the adaptive cascade mechanism:

\begin{itemize}
    \item \textbf{Face-based path:} $(27.53 + 18.82 + 3.67 + 1.45) \approx 51.47$ ms
    \item \textbf{Visual-only path:} $(34.54 + 18.82 + 3.67 + 1.45) \approx 58.48$ ms
    \item \textbf{Semantic-enhanced path:} $(2125.30 + 18.82 + 3.67 + 1.45) \approx 2149.24$ ms
\end{itemize}

Although semantic extraction requires approximately 2.1 s, it is activated for only 31.6\% of Market-1501 queries and 39.3\% of DukeMTMC-reID queries through the proposed adaptive cascade according to Table~\ref{tab:cascade_distribution}. Consequently, most queries are resolved using the lightweight visual branch, substantially reducing average inference latency compared with an always-on multimodal pipeline. This design makes the framework practical for latency sensitive surveillance applications while preserving semantic reasoning for difficult cases.

\begin{table}[H]
\centering
\caption{Component-wise Latency Analysis}
\label{tab:latency}
\small
\renewcommand{\arraystretch}{1.05}

\begin{tabular*}{\columnwidth}{@{\extracolsep{\fill}}lcccc}
\hline
\textbf{Component} 
& \multicolumn{3}{c}{\textbf{Latency (ms)}} 
& \textbf{\% Time} \\
\cline{2-4}
 & \textbf{Mean} & \textbf{Min} & \textbf{Max} & \\
\hline

Semantic Extraction & 2125.30 & 1455.04 & 3070.61 & 96.1 \\
Visual Extraction   & 34.54   & 19.90   & 71.96   & 1.6 \\
Face Extraction     & 27.53   & 2.19    & 283.08  & 1.2 \\
Detection + Crop    & 18.82   & 8.52    & 985.65  & 1.0 \\
Cascade Retrieval   & 3.67    & 0.67    & 86.54   & 0.1 \\
Vector DB Update    & 1.45    & 0.69    & 3.03    & 0.0 \\
\hline

\end{tabular*}
\end{table}

\begin{table}[H]
\centering
\caption{Cascade Path Distribution Across Datasets}
\label{tab:cascade_distribution}
\small
\renewcommand{\arraystretch}{1.05}

\begin{tabular*}{\columnwidth}{@{\extracolsep{\fill}}lcc|cc}
\hline
\textbf{Cascade Outcome} 
& \multicolumn{2}{c|}{\textbf{DukeMTMC-reID}} 
& \multicolumn{2}{c}{\textbf{Market-1501}} \\
\cline{2-5}

& \textbf{Queries} & \textbf{\%} 
& \textbf{Queries} & \textbf{\%} \\
\hline

Early-Exit & 1352 & 60.7 & 2303 & 68.4 \\
Escalated  & 876  & 39.3 & 1065 & 31.6 \\
No Match   & 0    & 0.0  & 0    & 0.0 \\
\hline

Total & 2228 & 100.0 & 3368 & 100.0 \\
\hline

\end{tabular*}
\end{table}

Table~\ref{tab:cascade_distribution} shows the cascade path distribution across both datasets. The results indicate that 60.7\% of DukeMTMC-reID queries and 68.4\% of Market-1501 queries are resolved through early-exit without semantic enhancement. Based on the observed cascade distribution, the estimated average query latency on DukeMTMC-reID is approximately 900\,ms, significantly lower than invoking semantic reasoning for every query.

To evaluate the robustness of the proposed adaptive cascade, we further conducted a threshold sensitivity analysis by varying the early-exit and margin thresholds. The results show that retrieval performance remains highly stable across different threshold configurations, with DukeMTMC-reID Rank-1 accuracy ranging from 90.1\% to 90.6\% and mAP from 81.6\% to 82.2\%, while Market-1501 Rank-1 accuracy ranges from 95.0\% to 95.3\% and mAP from 88.7\% to 89.1\%. These findings demonstrate that the proposed thresholding strategy primarily influences computational efficiency while preserving retrieval performance, validating the effectiveness of the adaptive early-exit cascade. Accordingly, the operating thresholds were set to \texttt{EARLY\_EXIT\_THRESHOLD}=0.75, \texttt{MARGIN\_THRESHOLD}=0.01, \texttt{STD\_THRESHOLD}=0.15, and \texttt{STAGE\_1\_TOP\_K}=20 for all reported experiments.

\subsection{\textbf{Qualitative Analysis}}

\begin{figure}[H]
    \centering
    \includegraphics[width=0.20\columnwidth]{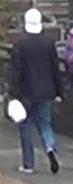}
    \hspace{0.02\columnwidth}
    \includegraphics[width=0.20\columnwidth]{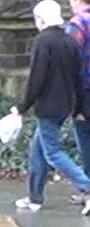}
    
    \caption{Qualitative comparison of the same identity across different camera views: (a) and (b).}
    \label{fig:qualitative_example}
\end{figure}

Fig.~\ref{fig:qualitative_example}(a) and Fig.~\ref{fig:qualitative_example}(b) show two samples of the same identity captured under different viewpoints.

For image (a), the semantic description is \textit{``dark gray hooded sweatshirt, solid color, long sleeves, dark indigo straight-leg pant''} 

For image (b), the semantic description is \textit{``dark gray long-sleeve jacket, solid color, full-length sleeves, dark blue straight-leg pants''}.

Although there are slight variations in wording (e.g., sweatshirt vs.\ jacket), both descriptions consistently capture key identity level attributes such as color, clothing type, and overall appearance. 

From a quantitative perspective, image (a) achieves a visual similarity score of 0.80 and a semantic similarity score of 0.97, resulting in a high overall matching score of 0.86. Image (b) shows a slightly lower visual similarity (0.78), but maintains a strong semantic similarity of 0.91, yielding an overall score of 0.83. This demonstrates that semantic representations provide more stable and discriminative features compared to visual embeddings alone. In both cases, the higher semantic scores compensate for moderate visual similarity, leading to correct identification.

\section{\textbf{Ablation Studies}}

\subsection{\textbf{Visual Feature Extractor}}

While CLIP provides robust, high-level semantic representations derived from large scale vision-language pre-training, it lacks the granular sensitivity required for precise person re-identification. We observed that CLIP’s latent space often collapses distinct identities into similar semantic clusters (e.g., 'person in dark jacket'), failing to distinguish between subtle identity specific cues. To address this, we transitioned to OSNet x1.0 (Omni Scale Network).

Unlike CLIP, OSNet x1.0 is explicitly optimized for the Re-ID task, utilizing a multi scale architecture to decouple identity from environmental noise such as viewpoint variations, lighting shifts, and occlusions. Using OSNet x1.0's ability to extract discriminative features, the system achieved a significant boost in performance. Furthermore, our analysis suggests that performance ceilings can be further elevated by integrating even more advanced Re-ID embedding architectures.

\subsection{\textbf{Semantic Feature Extractor}}

Beyond low-level visual embeddings, we evaluated the impact of incorporating high-level semantic descriptors generated by VLMs. This semantic branch provides complementary identity cues, particularly in scenarios with poor facial visibility or low sensor resolution. To maximize identity-specific information, we employed a structured prompt optimization strategy that captures:

\begin{enumerate}
\item \textbf{Upper Body:} Color, garment type, texture, and layering.
\item \textbf{Lower Body:} Garment type, fit, and length.
\item \textbf{Footwear:} Color and structural characteristics.
\item \textbf{Accessories:} Bags, eyewear, and headwear.
\item \textbf{Distinctive Traits:} Logos, stains, and other unique appearance cues.
\end{enumerate}

\subsection{\textbf{Inference Efficiency Analysis}}
As shown in Table \ref{tab:model_latency}, we benchmarked several state of the art VLMs to balance the trade off between descriptive richness and computational overhead. While the Gemma family provided detailed descriptors, it incurred significant latency, with gemma-4-E2B-it reaching 34.23s. In contrast, the Qwen series demonstrated superior efficiency; specifically, Qwen3-VL-4B achieved the lowest latency at 2.12s while maintaining high discriminative accuracy. This suggests that the Qwen3 architecture is particularly well-suited for near real time semantic feature extraction in person re-identification pipelines.

\begin{table}[t]
\centering
\caption{Semantic Inference Time}
\label{tab:model_latency}
\small
\renewcommand{\arraystretch}{1.05}

\begin{tabular*}{\columnwidth}{@{\extracolsep{\fill}}lc}
\hline
\textbf{Model Identifier} & \textbf{Latency (ms)} \\
\hline

google/gemma-3-4b-it  & 17,458.91 \\
google/gemma-4-E2B-it & 34,234.81 \\
Qwen2.5-VL-3B         & 6,624.44  \\
Qwen3-VL-4B           & 2,125.30  \\
\hline

\end{tabular*}
\end{table}




\section{\textbf{Conclusion}}

In this paper, we proposed a generative AI integrated multimodal framework for multi-camera person re-identification that balances retrieval accuracy and computational efficiency through an adaptive early-exit cascade.

Experimental results demonstrate that the proposed adaptive early-exit cascade resolves over 60\% of queries without requiring semantic reasoning, resulting in an average query latency of approximately 900 ms. These findings validate the effectiveness of selectively activating computationally expensive modalities only for ambiguous queries while maintaining competitive retrieval performance.

Overall, the proposed framework provides an efficient and scalable solution for real time multi-camera person re-identification through adaptive multimodal inference. By selectively activating computationally expensive modalities only when necessary, the framework effectively reduces inference latency while preserving retrieval performance, making it well suited for latency sensitive surveillance applications. Future work will focus on large scale real world evaluation, improving semantic inference efficiency, and integrating emerging foundation and transformer-based ReID models to further enhance both retrieval accuracy and computational efficiency.

\section{\textbf{Acknowledgment}}

The authorship team would like to acknowledge the vision, support and guidance of the IEEE Industrial Electronics Society in conducting the Generative AI Hackathon under the leadership of Daswin De Silva and Lakshitha Gunasekara.

\bibliographystyle{ieeetr}
\bibliography{sample}

@inproceedings{clipreid,
  author    = {S. Li and L. Sun and Q. Li},
  title     = {{CLIP-ReID: Exploiting Vision-Language Model for Image Re-Identification Without Concrete Text Labels}},
  booktitle = {Proceedings of the AAAI Conference on Artificial Intelligence},
  year      = {2023}
}

@inproceedings{domaingap,
  author    = {L. Wei and S. Zhang and W. Gao and Q. Tian},
  title     = {{Person Transfer GAN to Bridge Domain Gap for Person Re-Identification}},
  booktitle = {Proceedings of the IEEE/CVF Conference on Computer Vision and Pattern Recognition (CVPR)},
  pages     = {79--88},
  year      = {2018}
}

@article{wang2025visualanalysis,
  author  = {A. Asperti and L. Naldi and S. Fiorilla},
  title   = {{An Investigation of the Domain Gap in CLIP-Based Person Re-Identification}},
  journal = {Sensors},
  volume  = {25},
  number  = {2},
  pages   = {363},
  year    = {2025}
}

@article{clipscgi,
  author  = {Q. Han and X. He and Z. Liu and S. Liu and Y. Zhang and J. Xiang},
  title   = {{CLIP-SCGI: Synthesized Caption-Guided Inversion for Person Re-Identification}},
  journal = {arXiv preprint arXiv:2410.09382},
  year    = {2024}
}

@article{lvlmreid,
  author  = {Q. Wang and B. Li and X. Xue},
  title   = {{When Large Vision-Language Models Meet Person Re-Identification}},
  journal = {arXiv preprint arXiv:2411.18111},
  year    = {2024}
}

@article{videoreid,
  author  = {R. S. M. Saad and M. M. Moussa and N. S. Abdel-Kader and H. Farouk and S. Mashaly},
  title   = {{Deep Video-Based Person Re-Identification (Deep Vid-ReID): Comprehensive Survey}},
  journal = {EURASIP Journal on Advances in Signal Processing},
  volume  = {2024},
  number  = {63},
  year    = {2024}
}

@inproceedings{osnet,
  author    = {K. Zhou and Y. Yang and A. Cavallaro and T. Xiang},
  title     = {{Omni-Scale Feature Learning for Person Re-Identification}},
  booktitle = {Proc. IEEE/CVF Int. Conf. Comput. Vis. (ICCV)},
  pages     = {3702--3712},
  year      = {2019}
}

@inproceedings{facenet,
  author    = {F. Schroff and D. Kalenichenko and J. Philbin},
  title     = {{FaceNet: A Unified Embedding for Face Recognition and Clustering}},
  booktitle = {Proc. IEEE Conf. Comput. Vis. Pattern Recognit. (CVPR)},
  pages     = {815--823},
  year      = {2015}
}

@inproceedings{yolo,
  author    = {J. Redmon and S. Divvala and R. Girshick and A. Farhadi},
  title     = {{You Only Look Once: Unified, Real-Time Object Detection}},
  booktitle = {Proc. IEEE Conf. Comput. Vis. Pattern Recognit. (CVPR)},
  pages     = {779--788},
  year      = {2016}
}

@inproceedings{fasterreid,
  author    = {Y. Chen et al.},
  title     = {{Faster Person Re-Identification}},
  booktitle = {Proc. AAAI Conf. Artif. Intell.},
  year      = {2021}
}

@misc{fastreid,
  author = {Y. He et al.},
  title  = {{FastReID: A Pytorch Toolbox for General Instance Re-Identification}},
  year   = {2020},
  note   = {Available: https://github.com/JDAI-CV/fast-reid}
}

@article{mtcnn,
  author  = {K. Zhang and Z. Zhang and Z. Li and Y. Qiao},
  title   = {{Joint Face Detection and Alignment Using Multitask Cascaded Convolutional Networks}},
  journal = {IEEE Signal Processing Letters},
  volume  = {23},
  number  = {10},
  pages   = {1499--1503},
  year    = {2016}
}

@inproceedings{clip,
  author    = {A. Radford and J. W. Kim and C. Hallacy and A. Ramesh and G. Goh and S. Agarwal and G. Sastry and A. Askell and P. Mishkin and J. Clark and G. Krueger and I. Sutskever},
  title     = {{Learning Transferable Visual Models From Natural Language Supervision}},
  booktitle = {Proceedings of the 38th International Conference on Machine Learning (ICML)},
  pages     = {8748--8763},
  year      = {2021}
}

@inproceedings{transreid,
  author    = {S. He and H. Luo and P. Wang and F. Wang and H. Li and W. Jiang},
  title     = {{TransReID: Transformer-Based Object Re-Identification}},
  booktitle = {Proceedings of the IEEE/CVF International Conference on Computer Vision (ICCV)},
  pages     = {15013--15022},
  year      = {2021}
}

@inproceedings{duke,
  author    = {Ergys Ristani and Francesco Solera and Roger Zou and Rita Cucchiara and Carlo Tomasi},
  title     = {Performance Measures and a Data Set for Multi-Target, Multi-Camera Tracking},
  booktitle = {Proceedings of the European Conference on Computer Vision (ECCV) Workshops},
  pages     = {17--35},
  year      = {2016}
}

@inproceedings{market,
  author    = {Liang Zheng and Liyue Shen and Lu Tian and Shengjin Wang and Jingdong Wang and Qi Tian},
  title     = {Scalable Person Re-Identification: A Benchmark},
  booktitle = {Proceedings of the IEEE International Conference on Computer Vision (ICCV)},
  pages     = {1116--1124},
  year      = {2015}
}

@inproceedings{mgn,
  author    = {Guanshuo Wang and Yufeng Yuan and Xiong Chen and Jiwei Li and Xi Zhou},
  title     = {Learning Discriminative Features with Multiple Granularities for Person Re-Identification},
  booktitle = {Proceedings of the ACM International Conference on Multimedia (ACM MM)},
  pages     = {274--282},
  year      = {2018}
}

@inproceedings{dgnet,
  author    = {Zhedong Zheng and Xiaodong Yang and Zhiding Yu and Liang Zheng and Yi Yang and Jan Kautz},
  title     = {Joint Discriminative and Generative Learning for Person Re-Identification},
  booktitle = {Proceedings of the IEEE Conference on Computer Vision and Pattern Recognition (CVPR)},
  pages     = {2138--2147},
  year      = {2019}
}

@inproceedings{san,
  author    = {Xin Jin and Cuiling Lan and Wenjun Zeng and Guoqiang Wei and Zhibo Chen},
  title     = {Semantics-Aligned Representation Learning for Person Re-Identification},
  booktitle = {Proceedings of the AAAI Conference on Artificial Intelligence},
  pages     = {11173--11180},
  year      = {2020}
}

@inproceedings{pyramid,
  author    = {Feng Zheng and Cheng Deng and Xing Sun and Xinyang Jiang and Xiaowei Guo and Zongqiao Yu and Feiyue Huang and Rongrong Ji},
  title     = {Pyramidal Person Re-Identification via Multi-Loss Dynamic Training},
  booktitle = {Proceedings of the IEEE Conference on Computer Vision and Pattern Recognition (CVPR)},
  pages     = {8514--8522},
  year      = {2019}
}

@inproceedings{pfd,
  author    = {Tao Wang and Hong Liu and Pinhao Song and Tianyu Guo and Wei Shi},
  title     = {Pose-Guided Feature Disentangling for Occluded Person Re-Identification},
  booktitle = {Proceedings of the AAAI Conference on Artificial Intelligence},
  pages     = {2540--2549},
  year      = {2022}
}
\end{document}